\documentclass{ecai} 

\usepackage{latexsym}
\usepackage{amssymb}
\usepackage{amsmath}
\usepackage{amsthm}
\usepackage{booktabs}
\usepackage{enumitem}
\usepackage{graphicx}
\usepackage{color}
\usepackage{float}

\DeclareMathSymbol{\shortminus}{\mathbin}{AMSa}{"39}

\newcommand{\BibTeX}{B\kern-.05em{\sc i\kern-.025em b}\kern-.08em\TeX}

\begin{document}


\begin{frontmatter}


\paperid{123} 


\title{Enhancing Manufacturing Knowledge Access with LLMs and Context-aware Prompting}


\author[A]{\fnms{Sebastian}~\snm{Monka}}
\author[A]{\fnms{Irlan}~\snm{Grangel-Gonz\'alez}}
\author[A]{\fnms{Stefan}~\snm{Schmid}}
\author[A]{\fnms{Lavdim}~\snm{Halilaj}}
\author[B]{\fnms{Marc}~\snm{Rickart}}
\author[B]{\fnms{Oliver}~\snm{Rudolph}}
\author[B]{\fnms{Rui}~\snm{Dias}}

\address[A]{Bosch Center for Artificial Intelligence, Renningen, Germany\\
 {sebastian.monka, irlan.grangelgonzalez, stefan.schmid, lavdim.halilaj}@de.bosch.com}
\address[B]{Robert Bosch GmbH, Mobility Electronics, Reutlingen, Germany\\
 {marc.rickart, oliver.rudolph, rui.dias}@de.bosch.com}


\begin{abstract}
Knowledge graphs (KGs) have transformed data management within the manufacturing industry, offering effective means for integrating disparate data sources through shared and structured conceptual schemas.
However, harnessing the power of KGs can be daunting for non-experts, as it often requires formulating complex SPARQL queries to retrieve specific information.
With the advent of Large Language Models (LLMs), there is a growing potential to automatically translate natural language queries into the SPARQL format, thus bridging the gap between user-friendly interfaces and the sophisticated architecture of KGs.
The challenge remains in adequately informing LLMs about the relevant context and structure of domain-specific KGs, e.g., in manufacturing, to improve the accuracy of generated queries.
In this paper, we evaluate multiple strategies that use LLMs as mediators to facilitate information retrieval from KGs. 
We focus on the manufacturing domain, particularly on the Bosch Line Information System KG and the I40 Core Information Model.
In our evaluation, we compare various approaches for feeding relevant context from the KG to the LLM and analyze their proficiency in transforming real-world questions into SPARQL queries.
Our findings show that LLMs can significantly improve their performance on generating correct and complete queries when provided only the adequate context of the KG schema. 
Such context-aware prompting techniques help LLMs to focus on the relevant parts of the ontology and reduce the risk of hallucination. 
We anticipate that the proposed techniques help LLMs to democratize access to complex data repositories and empower informed decision-making in manufacturing settings. 
\end{abstract}

\end{frontmatter}


\section{Introduction}

Knowledge Graphs are essential in data management, especially for connecting complex data across different domains.
They are particularly valuable in manufacturing for integrating data from processes, machinery, and products.
KGs excel at consolidating data silos into structured, industry-relevant knowledge, which is crucial for Industry 4.0 applications focused on data-driven decision-making.
Alongside KGs, advancements in LLMs have transformed Natural Language Processing (NLP).
LLMs are not just adept at mimicking human language, but can also contextualize and provide insights by linking to KGs. This synergy enables intuitive user interaction with complex data, widening access to KGs for non-experts.
Although integrating LLMs with KGs for enhanced data interaction has attracted significant attention~\cite{pan-23,llm_kg_2024,rangel2024,Yang_24}, its application in the manufacturing domain remains in its infancy status.
Preliminary studies are promising~\cite{RonyKTK022,Zhu-24}, suggesting that LLMs could considerably elevate KG usability.
Such usability encompasses streamlined search features, the generation of insights from complex data relations, and conjectures of solutions built upon the KGs' data.
Despite these advancements, challenges persist, particularly in exploiting these capabilities to fulfill specific industrial objectives effectively.
The main reason for that remains the difficulty for non-experts to interact and query KGs, particularly to master query languages like SPARQL.
In this paper, we delve into the synergy of LLMs and KGs, particularly examining their potential to refine querying structured datasets via conversational language.
We study a repertoire of approaches addressing two key elements: 1) relaying KG data to LLMs and; and 2) converting conversational queries into structured SPARQL queries that can interact with the KG.
In our analysis, we introduce a specialized benchmark derived from the Line Information System (LIS) KG, which is reflective of the manufacturing sector's complexities~\cite{grangel-gonzalezLISKnowledgeGraphBased2023}.
The construct of this benchmark facilitates a detailed evaluation of diverse techniques for leveraging LLMs in KG query generation.
Our primary objective lies in elucidating insights and formulating best practices for utilizing LLMs as navigational tools to access KG information.
Our investigation underscores the transformative capabilities of LLMs—serving as gateways to knowledge and tools for empowerment within the data-dense landscape of manufacturing KGs. 

\begin{figure*}[htbp]
    \includegraphics[width=\linewidth]{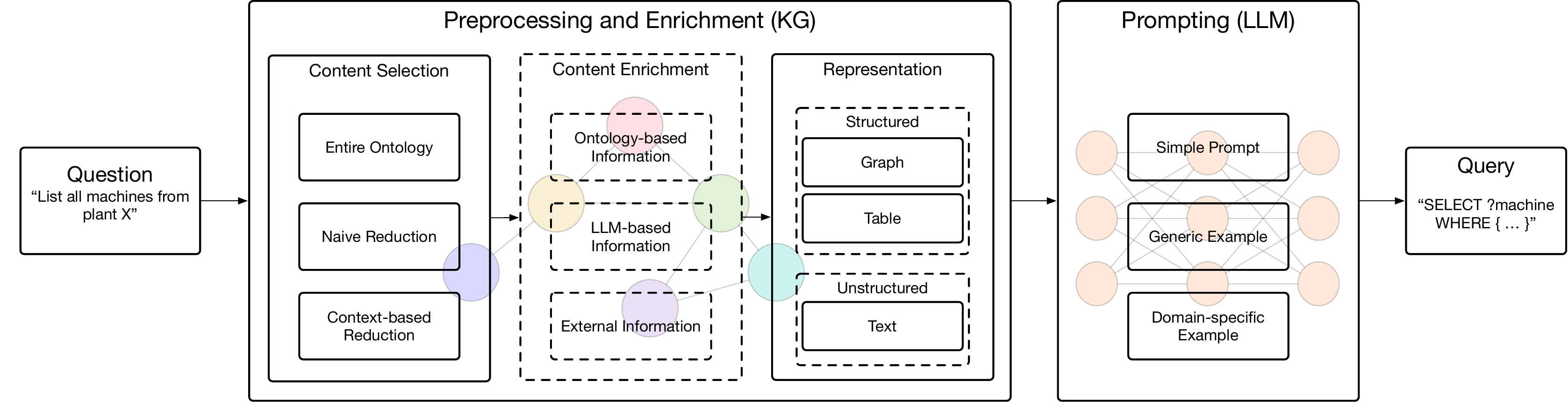}
    \caption{\textbf{LLM-based Knowledge Access Framework. It comprises two main steps: 1) \emph{Preprocessing and Enrichment (KG)} - providing relevant ontological content based on the given prompt; and 2) \emph{Prompting (LLM)} - enriching prompt with additional constraints and templates for guiding the LLM. 
    Each step contains dedicated components that are dynamically invoked based on the scenario.}}
    \label{fig:Framework}
\end{figure*}

\section{Related Work}
\label{sec:related-work}

The combination of KGs and LLMs has sparked considerable interest in the actual discussions.
Pan \emph{et al.}~\cite{llm_kg_2024} surveyed several recently proposed approaches and presented a roadmap about how LLM and KG could be unified.
They provided three different architectures, namely, KG-enhanced LLMs, LLM-augmented KGs, and Synergized LLMs + KGs.
Stevens \emph{et al.}~\cite{stevensAutomatingGenerationTextual2011} proposed a method to automatically generate text-based definitions from an ontology with logical descriptions of its entities, while~\cite{e.vOntologyVerbalizationUsing2016} followed a rule-based technique to create redundancy-free natural language descriptions from RDF.
A variety of works have targeted the problem of converting natural language to SPARQL.
Natural language triples either can be directly mapped to RDF triples~\cite{kaufmannQuerixNaturalLanguage} or a neural network can be trained to transform the text directly to SPARQL~\cite{luzSemanticParsingNatural2018, soruSPARQLForeignLanguage}.
Yang \emph{et al.}~\cite{Yang_24} presented an approach to generate SPARQL with selected schemas. 
Rangel \emph{et al.}~\cite{rangel2024} outlined a study that evaluates different strategies for fine-tuning OpenLlama LLM~\cite{openlm2023openllama}.
They investigated the SPARQL query generation on top of a domain-specific KG from the life science domain. 
Avila \emph{et al.}~\cite{10475614} reported on different experiments to transform natural language into SPARQL using GPT3.5. 
Lehman \emph{et al.}~\cite{ecai-Lehman23} described the use of controlled natural language for the generation of SPARQL queries.
In this work, authors pretrained LLMs on a large corpus of text to process natural language. 
Yuan \emph{et al.}~\cite{Yuan-24} studied the combination of user questions and employed GPT3.5 to generate SPARQL queries.
Here, a domain KG is used, i.e., the Metal-Organic Frameworks KG. 
Rony \emph{et al.}~\cite{RonyKTK022} pinpointed current issues with template-based options to generate SPARQL queries using LLMs.
Within their approach, named SGPT, authors employed embedding techniques and training techniques to enable the system to apprehend patterns and also use graph information to enhance the query generation process by the LLM. 
Taffa \emph{et al.}~\cite{TaffaU23} developed an approach that utilizes GPT3.5 to generate SPARQL queries with a focus on the scholarly domain. 

Previous research can only be considered a first step towards a profound understanding of SPARQL query generation from natural language questions using LLMs.  
Our study in this paper aims to delve deeper into the performance of LLMs in this task. We explore ways to enhance their performance through novel context-aware content selection and enrichment techniques. 
Furthermore, we concentrate on addressing the complexities of natural language-based knowledge access within the manufacturing sector.

\section{LLM-based Knowledge Access}
Within the manufacturing industry, LLMs are seen as promising enablers for leveraging data captured and represented by KGs~\cite{ZHOU2024102333}.
Their ability to provide intuitive access to complex KGs through natural language offers considerable benefits, particularly for users without specialized expertise.
To ensure that the LLM can generate correct SPARQL based on domain-specific semantic details it needs to be informed about knowledge in the KG.
Critical to this endeavor is the bidirectional translation of data formats.
Firstly, converting the KG triples into a format that the LLM can understand.
Secondly, translating user-expressed natural language queries into the SPARQL format for KG querying.

\subsection{Preprocessing and Enrichment}
\label{ssec:Preprocessing and Enrichment}
LLMs have demonstrated the promising ability to convert text-based questions into SPARQL queries that are syntactically correct.
Yet, the resulting queries often lack the specificity required for particular domains since LLMs may not have been exposed to domain-specific knowledge during their training.
To overcome this, LLMs must be equipped with appropriate information about the domain in question, which we describe as the process of content selection.
Once the LLM is provided with pertinent domain knowledge, the next challenge is to transform the selected triple-based structure of the KG into a representation that the LLM can interpret in a textual context.
This representation transformation is crucial as it ensures that the LLM can process the knowledge encoded in the KG and use it to inform its generation of SPARQL queries.


\subsubsection{Content Selection}
In aligning the general capabilities of LLMs to more specific domains and tasks, it is essential to provide contextually relevant content that enables these general models to access and apply domain-specific knowledge accurately.
This could be achieved through various means such as direct incorporation of the entire ontology, selection of sub-ontologies by naive reduction, or context-based reduction methods similar to basic Retrieval-Augmented Generation (RAG).

\paragraph{Entire Ontology:} 
An ontology equips LLMs with a higher-level comprehension of the KG, allowing for a more sophisticated understanding of how data is interconnected and can be queried.
Providing the ontology, allows LLMs to more effectively navigate the KG, discerning the structure and semantics of stored data which in turn enhances their ability to perform knowledge-intensive tasks.

\paragraph{Naive Reduction:} Ontologies can sometimes be expansive or cluttered, containing a wealth of classes and properties that may not all be pertinent to a given task.
In such cases, creating a sub-ontology that encapsulates a targeted subset of the broader schema can significantly assist the LLM.
This helps to manage the constraints posed by the token length limitations of the models.
Further, it can also diminish potential confusion that may arise from the overlapping concepts.
Encoding a carefully selected sub-ontology can improve the LLM's focus and increase its operational efficiency by streamlining only the domain context it needs to process.

\paragraph{Context-based Reduction:} Context-based reduction is a strategy to endow LLMs with only question-relevant knowledge from a KG.
For instance, when the LLM is presented with a question concerning the machinery in a factory, it would be primed with information specifically about industrial equipment and plant operations.
To facilitate this, a preprocessing step is required, extracting and structuring knowledge so that it can be fed into the LLM as a supplementary context.
A RAG technique is employed to identify the appropriate ontology concepts to the query through vector space similarity measures.
This additional layer enables the LLM to differentiate between relevant and irrelevant content for domain-specific inquiries.

\subsubsection{Content Enrichment}
\label{sssec:Content Enrichment}
Content enrichment serves as an optional step in augmenting the utility of the extracted information.
Three distinct variants can be distinguished, namely the enrichment with ontology-based information, LLM-based information, or external information.

\paragraph{Ontology-based Information:}
The ontology, with its inherent structure, can act as a resource to cultivate further explanatory content.
This is achieved by implementing heuristic rules to generate additional descriptions for classes and properties, thereby deepening the semantic richness of the ontology. 

\paragraph{LLM-based Information:}
Additionally, the LLM itself can contribute to the enrichment process. 
By inducing thought chains, the LLM can reflexively enhance and interlink the extracted content, weaving a tapestry of contextualized knowledge that is both expansive and coherent.

\paragraph{External Information:}
Moreover, external repositories can be scoured to integrate supplementary layers of meaning, extending beyond the ontology's confines. 
This could involve assimilating data from other knowledge bases or synthesizing structured or unstructured online information.

\subsubsection{Representation}
\label{sssec:Representattion}
The representational formats of KGs and LLMs occupy distinct positions along the spectrum of data structures.
KGs rely on a formal, structured arrangement of data, encapsulated in the form of triples, while LLMs excel in parsing and generating unstructured natural language.
Bridging these disparate models presents an important challenge in translating KG's structured context into a suitable format that LLMs can cope with.
Consequently, we are investigating various methodologies for presenting KG content to LLMs.

\paragraph{Graph Structure:} By nature, KGs are represented on a graph where nodes are connected by edges, each symbolizing various relationships.
However, this intrinsic graph format does not align seamlessly with LLMs' mode of operation, which may struggle to represent token sequences derived from URIs and triples efficiently.
Efficiently in this context refers to the token budget constraints encountered by LLMs when interpreting lengthy URIs.
Additionally, it's speculative whether the intricate network of node interrelations within a KG can be adequately comprehended by an LLM.
It is also essential to evaluate whether supplementary comments and descriptions within the KG significantly influence LLM performance.

\paragraph{Table Structure:} Adopting a table-based layout presents a more succinct way of representing KG data, albeit at the expense of losing the explicit depiction of graph relationships.
LangChain~\cite{Chase_LangChain_2022}, a notable framework in this domain, employs such a tabular structure alongside separate lists cataloging all concepts, such as classes, objects, and datatype properties.
The concepts defined in an ontology are encoded in the format of \textit{(<uri>, <uri>, <uri> or LITERAL)}.
However, LangChain omits the full URIs while only considering the local name and optionally the concept description in parentheses. 
Therefore, the semantic relations defined in an ontology such as inter-class or class-property relationships, respectively, are fully ignored.
In this configuration, metadata such as labels and comments can additionally serve to augment the informational content.

\paragraph{Text Structure:}
The final approach enlists an LLM to distill the ontology into a summarization, thus transforming the structured format into a narrative digestible by LLMs before initializing query generation.
This text-based transformation, however, bears the inherent risk of information attrition, with KG potentially being condensed in two sequential stages: first, by converting KG content into a textual synopsis, and subsequently, by translating this synopsis into SPARQL.
To address this, transformation rules that systematically unravel the graph structure into a sequence of natural language statements need to be developed.
It enables LLMs to comprehend the intricacies of the KG without needing to grapple with the graph's topology or traverse through multiple hops in the KG to gather information~\cite{fatemiTalkGraphEncoding2023}.

\subsection{Large Language Models}

As a result of the previous preprocessing and enrichment steps, the relevant content is selected according to the end user's natural language question and represented in a way that  LLMs can make optimal use of it.
This input is a prerequisite to instruct the LLM on generating SPARQL queries where the context is taken into consideration.

\subsubsection{Model Choice}
\label{sssec:model_choice}
The efficacy with which SPARQL is generated and understood by an LLM is greatly shaped by the underlying model choice.
In the following, we will describe two main choices w.r.t. model size, generality, and exposure to the amount of data.

\paragraph{Pretrained Model:} The capabilities of a pretrained model are largely determined by factors such as its parameter count, diversity, and the volume of training data, as well as the duration and thoroughness of its training regimen.
In general, models of larger size with exposure to SPARQL during training are predisposed to deliver superior performance.
The tokenization technique employed can also influence a model's interpretative accuracy for different representation formats of selected content, analogous to how GPT4's tokenizer redesign resulted in enhanced proficiency in handling tasks related to Python code structures in comparison to its predecessor, GPT3.5~\cite{openai2024gpt4}.

\paragraph{Finetuned Model:} Fine-tuning leverages a detailed dataset composed of question-SPARQL pairs, accustoming the LLM to the direct translation between natural language inquiries and SPARQL syntax.
The LLM, through this exposure, is trained to identify and assimilate patterns that delineate the query structure, learning to craft SPARQL statements in response to a broad array of natural language prompts.
Further refinement in SPARQL creation can be accomplished using reinforcement learning techniques supplemented with user feedback, wherein the model iteratively adjusts its approach based on successful query outcomes and validation through user interaction.

\subsubsection{Prompt Engineering}
The construction of the prompt significantly influences the quality of the produced query.
Next, we categorize the three strategies for incorporating additional KG content into the prompt as follows: 1) simple prompt; 2) generic example, and 3) domain-specific example.

\paragraph{Simple Prompt:} The simple prompt methodology combines the posed question, the represented KG content, and the directive to assemble a SPARQL query within the prompt.
This method relies on the LLM's intrinsic comprehension of what a SPARQL query entails, informed by the instructions included in the prompt that outline the desired format of the response.

\paragraph{Generic Example:} Enriching the prompt with a static example implies providing the model with a clearer indication of what a SPARQL query looks like.
This is particularly beneficial for LLMs that may not have been exposed to SPARQL during training, as the example enables them to infer the structure of such queries by drawing parallels with other data formats previously encountered in their learning process.

\paragraph{Domain-specific Example:} Given that the phrasing of SPARQL queries is intimately linked to the structuring of data within the ontology, context-based examples can be advantageous.
Such examples, incorporated alongside the simple prompt, illustrate question-SPARQL query pairs that are characteristic of the specific domain.
The expectation is that the LLM will extract the fundamental relationships between the questions and their corresponding SPARQL queries, leveraging the context to tailor queries more closely to the structure of the KG.

\section{Empirical Evaluation}
This empirical evaluation aims to investigate the competence of LLMs to access knowledge of KGs within the manufacturing domain.
To accomplish this, a manufacturing benchmark composed of business-relevant questions derived from real-world manufacturing scenarios is established.
This benchmark is crucial for assessing how well LLMs can process complex, domain-specific inquiries and transform natural language into structured queries via SPARQL.
Our study spans two ontologies, the Line Information System (LIS) and the Core Information Model (CIMM)~\cite{grangel-2020}, providing a holistic analysis of the LLMs' performance in varied schematic environments.
Herein, we focus on comparing different methodologies while providing a discourse on relevant content selection, format representations, and prompting strategies employed in the system.

\subsection{Manufacturing Benchmark}
The manufacturing benchmark, as depicted in Table~\ref{tab:lis_benchmark}, evaluates how well LLMs can write domain-specific SPARQL queries presented in natural language within a manufacturing context.
This benchmark encompasses a curated collection of 17 queries. 
The queries comprise the most important business questions from the following Personas, each reflecting various operational, analytical, and strategic queries.
The creation of Personas that work with the KG helps to better understand the requirements. 
It fosters a more user-centered solution targeting actual user needs.
In the following, we describe the five most interesting Personas for the domain, namely Benchmarking Engineer, Benchmarking Planner, Maintenance Planner, Technology Developer, and Data Engineer.
\begin{table*}[htbp]
    \centering
    \begin{tabular}{l|c|l}
        Persona & Question & SPARQL \\
        \hline
        Benchmarking engineer & How many Stations are on Line X from Plant Y? & ...\\
        OEM Benchmarking Planner & How many equipments are on Line Y in Plant X? & ...\\
        Maintenance Planner & Give me the list of material numbers for the line X? & ...\\
        Technology Developer & For process X: which other lines share that process within the same traceability database? & ...\\
        Data Engineer & For production line X: where and how can I access the transactional data for data analytics? & ...\\
    \end{tabular}
    \caption{\textbf{Manufacturing Benchmark}. Example questions of the Benchmark made by the different Personas.}
    \label{tab:lis_benchmark}
\end{table*}
To ensure a robust evaluation, each business question is paired with its corresponding SPARQL query (gold standard) for both ontologies.
The aim is to represent the precise formal request to retrieve the relevant information from the KG.

\subsubsection{Ontologies}
Both ontologies under study, the LIS and the CIMM, offer distinct perspectives on manufacturing data. 
The LIS ontology is used as the semantic schema in its correspondent KG and solution described here~\cite{grangel-gonzalezLISKnowledgeGraphBased2023}.
The LIS KG integrates a wide range of data from multiple manufacturing plants, lines, and machines. 
As of the time of evaluation, the LIS KG had incorporated data from more than 15 plants, over 2,700 production lines, 16,000 physical machines, and more than 400 distinct manufacturing processes—a testament to its scope and scalability.
These considerations span across the various data silos that exist in the form of Manufacturing Execution Systems, Enterprise Resource Planning systems, and Master Data systems.
The LIS ontology represents an abstraction layer over the underlying manufacturing data sources.
It functions as a use case ontology that was modeled according to the relational DB schema of the source data. 
The CIMM ontology is a community endeavor aiming to describe common concepts in manufacturing\footnote{https://github.com/eclipse-esmf/esmf-manufacturing-information-model}. 
It serves as a domain ontology and is based on the international IEC manufacturing standard. 
CIMM further extends the semantic expressiveness present in the LIS ontology. 
It serves as a rich extension of the well-known I40 Core Information Model.
In the remainder of this paper, we compare the performance of our pipeline for both ontology variants.

\subsection{Experiment}
In this section, we describe the criteria that we use on the manufacturing benchmark.

\subsubsection{Content Selection and Enrichment} \label{sssec:Content Selection}
As discussed in Section~\ref{ssec:Preprocessing and Enrichment}, the application of different preprocessing and enrichment steps influences which domain-specific content is selected and how it is transformed to be fed into the LLM.
We briefly explain how we select and represent the content of the ontology and how we feed the content into the LLM.

\paragraph{Naive reduction of entire ontology: [--> $Ont_A$]}  This variant applies \emph{naive reduction} on the \emph{entire ontology} to derive a smaller sub-ontology that fits into the context-window of the LLM. Specifically, we keep the following properties: \texttt{rdf:type}, \texttt{rdfs:label}, \texttt{rdfs:domain}, \texttt{rdfs:range}, \texttt{rdfs:subClassOf}, and \texttt{rdfs:subPropertyOf} to match the 32K token limit of our GPT4 models. 
        
\paragraph{Context-based reduction of $Ont_A$: [--> $Ont_B$]} This variant applies \emph{context-based reduction} on the \texttt{base} sub-ontology. In particular, for each natural language question, we find the 25 top-matching classes or properties in the ontology using the semantic similarity index (cf. Section~\ref{ssec:Preprocessing and Enrichment}). 
It includes only the question-relevant classes and properties and their neighboring concepts.  
Given that $Ont_B$ includes only the relevant concepts of $Ont_A$ for every question, it allows us to measure the impact of the context-based reduction on the LLM to focus and reduce hallucinations.
        
\paragraph{Context-based reduction of entire ontology: [--> $Ont_C$]} Likewise, this variant applies \emph{context-based reduction}, but on the \emph{entire ontology}. 
Unlike above, this variant includes the complete ontology definition for the matching classes and properties, including their description, i.e., \texttt{rdfs:comment} and axioms, e.g., \texttt{owl:inverseOf}. This variant allows us to evaluate to what extent LLMs are able to use rich ontology definitions in the process of generating accurate SPARQL queries.
        
\paragraph{Context-based reduction of the entire ontology plus ontology-based content enrichment: [--> $Ont_D$]} 
This variant builds on $Ont_C$ and enriches it with an additional post-processing step as shown in Section~\ref{sssec:Content Enrichment} using ontology-based enrichment. 

\subsubsection{Representation}
When the relevant content is selected it is necessary to define in which format it should be fed into the LLM.
In Section~\ref{sssec:Representattion} we discussed three specific representation formats for the use within the LLM prompt, namely structured graph and table, as well as unstructured text.
For the experiments in this paper, we focus on structured format representations. 
\textbf{Graph}: in the graph representation the RDF Turtle format is used. 
The graph is directly used to be fed into the LLM. \textbf{Table}: the table representation extracts all classes, object properties, and datatype properties and provides them in three independent lists.
This implementation is similar to the method $GraphSparqlQAChain$ in the LangChain framework.
\textbf{Table-Sorted}: similar to the table representation all the classes, object properties, and datatype properties are extracted.
However, in the next step, all the properties are assigned to the individual class.
We therefore refer to that as a table-sorted representation format of the ontology.

\begin{table*}[!ht]
\centering
\begin{tabular*}{\textwidth}{c @{\extracolsep{\fill}} c}
\begin{minipage}[t]{0.48\textwidth}
\begin{tabular}{l|c|c|c|c}
\textbf{Accuracy} & \textbf{$Ont_A$} & \textbf{$Ont_B$} & \textbf{$Ont_C$} & \textbf{$Ont_D$} \\
\hline
$P_{simple}$(graph) & - & \textbf{0.95} & 0.93 & 0.91  \\
$P_{simple}$(table) & 0.59 & 0.83 & \textbf{0.85} & 0.81 \\
$P_{simple}$(table-sorted) & 0.47 & 0.82 & \textbf{0.86} & \textbf{0.86}  \\
\hline
$P_{example}$(graph) & - & 0.92 & \textbf{0.96} & 0.91 \\
$P_{example}$(table) & 0.72 & \textbf{0.88} & 0.82 & 0.86 \\
$P_{example}$(table-sorted) & 0.78 & \textbf{0.86} & 0.84 & 0.83 \\
\hline
$P_{domain}$(graph) & - & 0.85 & 0.84 & \textbf{0.96} \\
$P_{domain}$(table) & 0.80 & 0.90 & \textbf{0.97} & \textbf{0.97} \\
$P_{domain}$(table-sorted) & 0.92 & 0.94 & \textbf{0.96} & \textbf{0.96} \\
\end{tabular}
\caption{\textbf{Comparison of model hallucination accuracy across different prompts on LIS using GPT3.5}.}
\label{tab:accuracies_lis}
\end{minipage}

&  

\begin{minipage}[t]{0.48\textwidth}
\begin{tabular}{l|c|c|c|c}
\textbf{Accuracy} & \textbf{$Ont_A$} & \textbf{$Ont_B$} & \textbf{$Ont_C$} & \textbf{$Ont_D$} \\
\hline
$P_{simple}$(graph) & - & \textbf{0.92} & 0.63 & 0.71 \\
$P_{simple}$(table) & 0.76 & 0.87 & \textbf{0.90} & 0.84 \\
$P_{simple}$(table-sorted) & 0.81 & \textbf{0.92} & \textbf{0.92} & 0.90 \\
\hline
$P_{example}$(graph) & - & \textbf{0.94} & 0.69 & 0.75 \\
$P_{example}$(table) & 0.84 & 0.93 & \textbf{0.95} & 0.93 \\
$P_{example}$(table-sorted) & 0.87 & 0.91 & \textbf{0.94} & 0.88 \\
\hline
$P_{domain}$(graph) & - & \textbf{0.95} & 0.88 & 0.81 \\
$P_{domain}$(table) & 0.85 & \textbf{0.96} & 0.92 & 0.93\\
$P_{domain}$(table-sorted) & 0.91 & 0.91 & \textbf{0.92} & 0.89\\
\end{tabular}
\caption{\textbf{Comparison of model hallucination accuracy across different prompts on CIMM using GPT3.5}.}
\label{tab:accuracies_cimm}
\end{minipage}

\end{tabular*}
\end{table*}

\subsubsection{Prompting}
Prompting has shown to be important to instruct LLMs to fulfill especially domain-specific tasks.
In our experiments, we evaluated the following prompting techniques:
\paragraph{Simple Prompt ($P_{simple}$):}
This version uses a standard prompt instruction to guide the LLM in the direction of SPARQL output.
``Write a SPARQL query to answer the following question:
\{question\}.
Use the following ontology as schema for your query:
\{ontology\}''.

\paragraph{Prompt with Example ($P_{example}$):}
In addition to the simple prompt, a generic example is added, such as:
``Task: Generate a SPARQL SELECT statement for querying a graph database.
For instance, to find all machines of a given plant, the following query would be suitable: \{query\}''.

\paragraph{Prompt with Domain-Specific Example ($P_{domain}$):}
For the domain-specific prompt we use the example prompt and add a domain-specific example SPARQL query of the used ontology, e.g.:
``EXAMPLE 2:
For instance, to find all Materials, i.e., their numbers, that are used on Line Y in Plant X: \{query\}''.

\subsection{Quantitative Evaluation}
In the scope of the quantitative evaluation, we concentrate on the precision of domain-specific terms within the SPARQL queries, particularly analyzing the tendency of the employed approach to hallucinate, i.e., generate incorrect or non-existent terms.
We undertake this by comparing the individual subjects $s$, predicates $p$, and objects $o$ in the resulting SPARQL queries to their counterpart entities within the original ontology.
Should the particular node or relationship be present in the ontology, it is considered a match; conversely, its absence leads to classification as a mismatch.
Consequently, for every experimental approach, we determine a corresponding hallucination accuracy based upon the standard set of questions, the number of matches $N_{matches}$, and the number of mismatches $N_{mismatches}$.

$$N_{\text{matches}} = \left| \{ ({s,p,o}) \in \text{SPARQL} \,|\, ({s,p,o}) \in \text{Ontology} \} \right|$$
$$N_{\text{mismatches}} = \left| \{ ({s,p,o}) \in \text{SPARQL} \,|\, ({s,p,o}) \notin \text{Ontology} \} \right|$$

The accuracy metric $Acc$ is defined as follows:
\[ Acc = \frac{N_{matches}}{(N_{matches}+N_{mismatches})} \]
Given the inherent non-determinism of LLMs, each experiment is replicated five times to mitigate variance in the results.
The final reported accuracy represents the mean value calculated across all individual experimental runs.
We therefore compare the different methods with the proposed framework based on their selected content, the representation format, and the prompting technique.

The hallucination accuracy provides insights into the model performance for domain-specific tasks.
As depicted in Figure~\ref{tab:accuracies_lis} the models' hallucination accuracy is influenced by the specific content selected from the LIS ontology.
For the experiments, we use GPT3.5-16K as LLM.
It can be seen that the $Ont_A$ ontology configuration is more susceptible to hallucinations, implying a potential misalignment with the domain-specific facts when generating the SPARQL query.
On the contrary, a context-based reduced ontology, e.g. $Ont_C$, is designed to streamline only preselected domain-specific information.
Therefore, it appears to aid in better information memorization and recall, as suggested by the observed increase in accuracy.
A parallel trend is observable in the performance on the CIMM ontology (cf. Figure~\ref{tab:accuracies_cimm}).
Here, the LLM accuracy depends on the specificity of the selected content concerning the posed question.
We think a well-constructed ontology that aligns closely with the query will likely facilitate a more precise extraction of true content, resulting in enhanced model performance.
However, this cannot be proven with just the hallucination accuracy.
Nonetheless, when deliberating on the different representation formats, e.g. graph, table, and table-sorted, it is difficult to extract a consistent pattern.
This suggests that the inherent structure of the ontology data does not singularly dictate the models' hallucination tendencies.
Moreover, the selection of prompting techniques does not showcase a significant impact on the hallucination rates. 
Whether a graph-based or a table-based prompt is utilized, the incidence of hallucination remains seemingly invariant, underscoring a broader challenge in content retrieval and representation that extends beyond the formatting paradigms.


\begin{table}[htbp]
    \centering
    \begin{tabular}{l|c|c|c|c}
        \textbf{Accuracy}  & \textbf{$Ont_A$} & \textbf{$Ont_B$} & \textbf{$Ont_C$} & \textbf{$Ont_D$} \\
         \hline
        GPT3.5 - $P_{example}$(graph) & - & 0.94 & 0.69 & 0.75\\
        GPT4 - $P_{example}$(graph) & 0.90 & \textbf{0.95} & 0.85 & 0.90 \\
        \hline
        GPT3.5 - $P_{domain}$(graph) & - & \textbf{0.95} & 0.88 & 0.81\\
        GPT4 - $P_{domain}$(graph) & \textbf{0.93} & \textbf{0.95} &\textbf{ 0.93} & \textbf{0.91} \\
    \end{tabular}
    \caption{\textbf{Comparison of GPT3.5 and GPT4 on CIMM using the hallucination accuracy.}}
    \label{tab:model_choice}
\end{table}

As introduced in Section~\ref{sssec:model_choice} the choice of the LLM highly influences its performance.
Figure~\ref{tab:model_choice} reflects this also with the hallucination metric for natural language to SPARQL query tasks.
Hereby, it can be seen that GPT4 outperforms GPT3.5 in all variants concerning how well domain-specific knowledge is represented.

\subsection{Qualitative Evaluation}

To complete the evaluation of the approach, we study the performance of LLMs on the task of generating accurate SPARQL queries from natural language questions. 
We use three highly qualified knowledge engineers and domain experts to rate the generated queries with respect to their \emph{correctness} and \emph{completeness}.
In our qualitative evaluation, we use GPT4, i.e., the best model according to our quantitative evaluation, with a context window of 32K tokens.   

\begin{figure}[htbp]
    \centering
    \includegraphics[width=70mm]{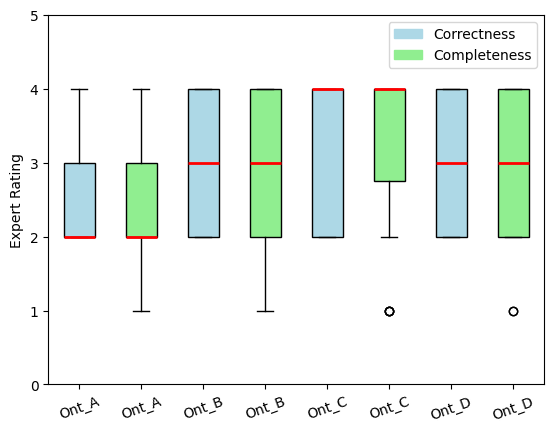}
    \caption{\textbf{Correctness and completeness evaluation results using GPT4-32K}. The figure depicts a box plot of the quantitative evaluation results of all generated SPARQL queries across the benchmark questions for the different content selection and enrichment variants using the prompt template with a generic example $P_{example}$.}
    \label{fig:boxplot}
\end{figure}

Concerning \emph{correctness}, the experts rated the generated SPARQL queries according to the following criteria: 0 (generation failure, i.e., too many tokens), 1 (syntactically erroneous, i.e., not conform to SPARQL specification), 2 (syntactically correct, but semantically wrong according to CIMM), 3 (syntactically and semantically correct - but not fully as intended by the ontology), 4 (syntactically and semantically correct and as intended by the ontology).
Concerning \emph{completeness}, the experts rated the generated queries.
They conducted a sequence of three assessments to evaluate the additional effort needed for query correction and refinement.
The rating schema is as follows: 0 (query generation failed --> $0\%$ completed), 1 (requires major changes --> less than $64\%$ completed), 2 (requires significant changes --> $65\shortminus89\%$ completed), 3 (requires minor changes --> $90\shortminus99\%$ completed), and 4 (query is complete and correct).
The average inter-rater reliability of our domain expert ratings across the different experiments and questions amounts to Fleiss' kappa values of $0.54$ for correctness and $0.29$ for completeness.
The results for the different variants using Prompt $P_{example}$ are illustrated in Figure~\ref{fig:boxplot}.

\begin{figure}[htbp]
    \centering
    \includegraphics[width=70mm]{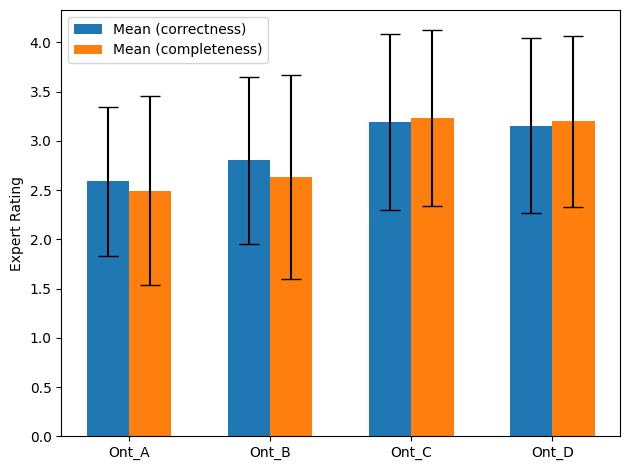}
    \caption{\textbf{Comparison of the quantitative evaluation results for the different variants using GPT4-32K}. The figure depicts the mean and standard deviation values for the different content selection and enrichment variants using the prompt template with a generic example $P_{example}$.}
    \label{fig:mean}
\end{figure}

The results across all queries clearly show that both, the correctness and the completeness increase when reducing the ontology to a contextualized sub-ontology that only includes the relevant concepts, i.e., classes and properties. 
The fact that the results for the variant $Ont_B$ outperform variant $Ont_A$ precisely states that the reduction of the ontology helps the LLM to focus on the relevant concepts. 
Furthermore, the performance improvement between variant $Ont_B$ and $Ont_C$ highlights that the LLMs can leverage the additional semantics provided by the rich ontology.
Since the baseline ontology for these experiments, i.e., CIMM has already rich definitions, e.g., \texttt{rdfs:comments} and axioms, e.g., \texttt{rdfs:domain} and \texttt{rdfs:range}, there exists no significant gain measurable as a result of content enrichment ($Ont_D$) in this case.


\begin{table}[htbp]
\centering
\begin{tabular}{l|lll|lll|}
\cline{2-7}
                                & \multicolumn{3}{c|}{\textbf{$P_{example}$}}                                                & \multicolumn{3}{c|}{\textbf{$P_{domain}$}}                                                \\ \cline{2-7} 
                                & \multicolumn{1}{l|}{\textbf{Mean}} & \multicolumn{1}{l|}{\textbf{Med}} & \textbf{Std} & \multicolumn{1}{l|}{\textbf{Mean}} & \multicolumn{1}{l|}{\textbf{Med}} & \textbf{Std} \\ \hline
\multicolumn{1}{|l|}{$Ont_A$}      & \multicolumn{1}{l|}{2.54}          & \multicolumn{1}{l|}{2.0}          & 0.71         & \multicolumn{1}{l|}{\textbf{2.70}}          & \multicolumn{1}{l|}{2.0}          & 0.87         \\ \hline
\multicolumn{1}{|l|}{$Ont_B$} & \multicolumn{1}{l|}{2.79}          & \multicolumn{1}{l|}{3.0}          & 0.82         & \multicolumn{1}{l|}{\textbf{2.82}}          & \multicolumn{1}{l|}{2.0}          & 0.93         \\ \hline
\multicolumn{1}{|l|}{$Ont_C$}   & \multicolumn{1}{l|}{3.14}          & \multicolumn{1}{l|}{3.0}          & 0.88         & \multicolumn{1}{l|}{\textbf{3.35}}          & \multicolumn{1}{l|}{4.0}          & 0.91         \\ \hline
\multicolumn{1}{|l|}{$Ont_D$}   & \multicolumn{1}{l|}{3.15}          & \multicolumn{1}{l|}{3.0}          & 0.80         & \multicolumn{1}{l|}{\textbf{3.18}}          & \multicolumn{1}{l|}{4.0}          & 0.96         \\ \hline
\end{tabular}
\caption{\textbf{Comparing the correctness ratings of the generated queries for different prompting approaches using GPT4-32K}: i.) with a generic example ($P_{example}$) and ii.) with a domain-specific example $P_{domain}$.}
    \label{tab:qual_table}
\end{table}

We further evaluated the correctness improvement of the generated queries when using a generic query example ($P_{example}$) vs. using a domain-specific query example ($P_{domain}$) in the prompt. 

\section{Discussion}
\label{sec:discussion}
The non-deterministic nature of the system presents a salient challenge in guaranteeing consistent performance during the generation of SPARQL queries.
As a result, there is a compelling need for the development of methods that either impose effective constraints on the generative process or furnish compensatory mechanisms to counteract the unpredictability.
Our study did not delve into the realm of fine-tuning, predominantly due to the scarcity of natural language to SPARQL query pairs, which are vital for such an endeavor.
It remains clear that the assembly of a comprehensive and domain-specific dataset for fine-tuning purposes constitutes a valuable avenue for future research.
While our quantitative evaluation centered on the hallucination score, it is crucial to acknowledge that this metric alone does not fully encapsulate the ability of a model to generate accurate SPARQL queries. Nevertheless, it gives a good insight into how well the LLMs have adapted to a particular domain, thus providing an indirect indicator of its potential performance.
It is core to highlight the complexity of the manufacturing domain and the LIS KG. 
The CIMM ontology, whose intent is to describe the domain, is built based on an international standard that models the physical and the logical world in different conceptual models. 
It is even for human experts often a challenging task to navigate such big ontologies and write correct SPARQL queries. For example, the concept \texttt{cimm:Equipment} in the CIMM ontology is used to describe logical concepts, e.g., Plant, Lines, Stations. However, in reality, domain experts often use the term equipment to refer to physical assets, as machines. This implies that to write a SPARQL query requesting all machines in a given plant, these nuances of the domain have to be properly understood. 
While running the experiments, we observed that despite the fact of semantic complexity our context-aware prompting strategies managed to handle the query generation quite well (cf. Table~\ref{tab:accuracies_cimm}).
The results of the qualitative evaluation consistently show that the correctness and the completeness ratings of the generated queries significantly increase by $23.2\%$ and $29.7\%$ respectively on average across all questions by reducing LLM input to the relevant concepts, i.e., approach $Ont_A$ vs. $Ont_C$. Furthermore, the performance improvement between variant $Ont_B$ and $Ont_C$ by $13.9\%$ (correctness) and $22.3\%$ (completeness) highlights that the LLMs can leverage the additional semantics provided by a rich ontology.

\section{Conclusion and Outlook}
\label{sec:lessons-learned-conclusions}
In this article, we have presented a novel LLM-based Knowledge Access approach, a framework that uses the power of LLMs, enhanced with domain-specific manufacturing data, to generate SPARQL queries from natural language questions about our Line Information System KG at Bosch.
We explore different context-aware prompting techniques that involve various content selection, enrichment, and representation approaches, as well as different templates for the LLM prompt.
We have performed extensive evaluation utilizing two domain ontologies for manufacturing, the Line Information System and the Core Information Model. 

Our results demonstrate that, in particular, context-aware content selection and enrichment reduced the risk for LLM hallucinations, and in general improved the accuracy of the LLM-generated SPARQL queries.
Across all our experiments we measured an average accuracy gain (across all benchmark queries) in the order of $20\shortminus30\%$ on our KG.
Moreover, our results indicate how different prompting techniques impact query accuracy and the risk of LLM hallucination. Specifically, we measured an accuracy improvement across all benchmark queries in the order of $5\shortminus8\%$ by including a single domain-specific example in the prompt compared to a generic example. 


As for the future work, we envision to handle more complex questions.
For instance, it should answer intricate questions such as, “List all lines that fulfill the requirements for the production of a given product”. 
This involves refining the LLM’s understanding of nuanced queries and its ability to generate corresponding SPARQL queries.
Furthermore, non-determinism and noise in answer generation need to be addressed.
Implementing ensemble generation and answer rating mechanisms could help to provide robust answers to user questions.
Another avenue of research we plan to investigate is the utilization of Open Language Models, e.g., Llama 3, Falcon and compare their performance to GPT-based LLMs for context-aware SPARQL generation.


\bibliography{mybibfile}

\begin{thebibliography}{22}
\providecommand{\natexlab}[1]{#1}
\providecommand{\url}[1]{\texttt{#1}}
\expandafter\ifx\csname urlstyle\endcsname\relax
  \providecommand{\doi}[1]{doi: #1}\else
  \providecommand{\doi}{doi: \begingroup \urlstyle{rm}\Url}\fi

\bibitem[An et~al.(2023)An, Greenberg, Kalinowski, Zhao, Hu, Uribe{-}Romo, Langlois, Furst, and G{\'{o}}mez{-}Gualdr{\'{o}}n]{Yuan-24}
Y.~An, J.~Greenberg, A.~Kalinowski, X.~Zhao, X.~Hu, F.~J. Uribe{-}Romo, K.~Langlois, J.~Furst, and D.~A. G{\'{o}}mez{-}Gualdr{\'{o}}n.
\newblock Knowledge graph question answering for materials science {(KGQA4MAT):} developing natural language interface for metal-organic frameworks knowledge graph {(MOF-KG)}.
\newblock \emph{CoRR}, abs/2309.11361, 2023.
\newblock \doi{10.48550/ARXIV.2309.11361}.
\newblock URL \url{https://doi.org/10.48550/arXiv.2309.11361}.

\bibitem[Avila et~al.(2024)Avila, Vidal, Franco, and Casanova]{10475614}
C.~V.~S. Avila, V.~M. Vidal, W.~Franco, and M.~A. Casanova.
\newblock Experiments with text-to-sparql based on chatgpt.
\newblock In \emph{IEEE 18th Int. Conf. on Semantic Computing (ICSC)}, pages 277--284, 2024.
\newblock \doi{10.1109/ICSC59802.2024.00050}.

\bibitem[Chase(2022)]{Chase_LangChain_2022}
H.~Chase.
\newblock {LangChain}, Oct. 2022.
\newblock URL \url{https://github.com/langchain-ai/langchain}.

\bibitem[E.~V and Kumar(2016)]{e.vOntologyVerbalizationUsing2016}
V.~E.~V and P.~S. Kumar.
\newblock Ontology {{Verbalization}} using {{Semantic-Refinement}}, Oct. 2016.

\bibitem[Fatemi et~al.(2023)Fatemi, Halcrow, and Perozzi]{fatemiTalkGraphEncoding2023}
B.~Fatemi, J.~Halcrow, and B.~Perozzi.
\newblock Talk like a {{Graph}}: {{Encoding Graphs}} for {{Large Language Models}}, Oct. 2023.

\bibitem[Geng and Liu(2023)]{openlm2023openllama}
X.~Geng and H.~Liu.
\newblock Openllama: An open reproduction of llama, May 2023.
\newblock URL \url{https://github.com/openlm-research/open_llama}.

\bibitem[Grangel{-}Gonz{\'{a}}lez et~al.(2023)Grangel{-}Gonz{\'{a}}lez, Rickart, Rudolph, and Shah]{grangel-gonzalezLISKnowledgeGraphBased2023}
I.~Grangel{-}Gonz{\'{a}}lez, M.~Rickart, O.~Rudolph, and F.~Shah.
\newblock {LIS:} {A} {K}nowledge {G}raph-{B}ased {L}ine {I}nformation {S}ystem.
\newblock In C.~Pesquita, E.~Jim{\'{e}}nez{-}Ruiz, J.~P. McCusker, D.~Faria, M.~Dragoni, A.~Dimou, R.~Troncy, and S.~Hertling, editors, \emph{The Semantic Web - 20th Int., {ESWC}, Hersonissos, Crete, Greece, May 28 - June 1, Proceedings}, volume 13870 of \emph{LNCS}, pages 591--608. Springer, 2023.

\bibitem[Grangel-González et~al.(2020)Grangel-González, Lösch, and ul~Mehdi]{grangel-2020}
I.~Grangel-González, F.~Lösch, and A.~ul~Mehdi.
\newblock Knowledge graphs for efficient integration and access of manufacturing data.
\newblock In \emph{25th IEEE Int. Conf. on Emerging Technologies and Factory Automation (ETFA)}, volume~1, pages 93--100, 2020.

\bibitem[Lehmann et~al.(2023)Lehmann, Ferr{\'{e}}, and Vahdati]{ecai-Lehman23}
J.~Lehmann, S.~Ferr{\'{e}}, and S.~Vahdati.
\newblock Language models as controlled natural language semantic parsers for knowledge graph question answering.
\newblock In K.~Gal, A.~Now{\'{e}}, G.~J. Nalepa, R.~Fairstein, and R.~Radulescu, editors, \emph{{ECAI} - 26th European Conference on Artificial Intelligence, September 30 - October 4, 2023, Krak{\'{o}}w, Poland - Including 12th Conference on Prestigious Applications of Intelligent Systems {(PAIS} 2023)}, volume 372 of \emph{Frontiers in Artificial Intelligence and Applications}, pages 1348--1356. {IOS} Press, 2023.
\newblock URL \url{https://doi.org/10.3233/FAIA230411}.

\bibitem[Luz and Finger(2018)]{luzSemanticParsingNatural2018}
F.~F. Luz and M.~Finger.
\newblock Semantic {{Parsing Natural Language}} into {{SPARQL}}: {{Improving Target Language Representation}} with {{Neural Attention}}, Mar. 2018.

\bibitem[OpenAI et~al.(2024)OpenAI, Achiam, Adler, Agarwal, Ahmad, Akkaya, Aleman, Almeida, and et~al.]{openai2024gpt4}
OpenAI, J.~Achiam, S.~Adler, S.~Agarwal, L.~Ahmad, I.~Akkaya, F.~L. Aleman, D.~Almeida, and J.~A. et~al.
\newblock Gpt-4 technical report, 2024.

\bibitem[Pan et~al.(2023)Pan, Luo, Wang, Chen, Wang, and Wu]{pan-23}
S.~Pan, L.~Luo, Y.~Wang, C.~Chen, J.~Wang, and X.~Wu.
\newblock Unifying large language models and knowledge graphs: {A} roadmap.
\newblock \emph{CoRR}, abs/2306.08302, 2023.
\newblock \doi{10.48550/ARXIV.2306.08302}.
\newblock URL \url{https://doi.org/10.48550/arXiv.2306.08302}.

\bibitem[Pan et~al.(2024)Pan, Luo, Wang, Chen, Wang, and Wu]{llm_kg_2024}
S.~Pan, L.~Luo, Y.~Wang, C.~Chen, J.~Wang, and X.~Wu.
\newblock Unifying large language models and knowledge graphs: A roadmap.
\newblock \emph{IEEE Transactions on Knowledge and Data Engineering (TKDE)}, 2024.

\bibitem[Reyes et~al.(2024)Reyes, de~Farias, Sima, and Kobayashi]{rangel2024}
J.~C.~R. Reyes, T.~M. de~Farias, A.~C. Sima, and N.~Kobayashi.
\newblock {SPARQL} generation: an analysis on fine-tuning {O}pen{LL}a{M}a for question answering over a life science knowledge graph.
\newblock \emph{CoRR}, abs/2402.04627, 2024.
\newblock \doi{10.48550/ARXIV.2402.04627}.
\newblock URL \url{https://doi.org/10.48550/arXiv.2402.04627}.

\bibitem[Rony et~al.(2022)Rony, Kumar, Teucher, Kovriguina, and Lehmann]{RonyKTK022}
M.~R. A.~H. Rony, U.~Kumar, R.~Teucher, L.~Kovriguina, and J.~Lehmann.
\newblock {SGPT:} {A} generative approach for {SPARQL} query generation from natural language questions.
\newblock \emph{{IEEE} Access}, 10:\penalty0 70712--70723, 2022.

\bibitem[Soru et~al.(2017)Soru, Marx, Moussallem, Publio, Valdestilhas, Esteves, and Neto]{soruSPARQLForeignLanguage}
T.~Soru, E.~Marx, D.~Moussallem, G.~Publio, A.~Valdestilhas, D.~Esteves, and C.~B. Neto.
\newblock {SPARQL} as a foreign language.
\newblock In J.~D. Fern{\'{a}}ndez and S.~Hellmann, editors, \emph{Proceedings of the Posters and Demos Track of the 13th International Conference on Semantic Systems - SEMANTiCS2017 co-located with the 13th International Conference on Semantic Systems (SEMANTiCS 2017), Amsterdam, The Netherlands, September 11-14, 2017}, volume 2044 of \emph{{CEUR} Workshop Proceedings}. CEUR-WS.org, 2017.
\newblock URL \url{https://ceur-ws.org/Vol-2044/paper14/}.

\bibitem[Stevens et~al.(2011)Stevens, Malone, Williams, Power, and Third]{stevensAutomatingGenerationTextual2011}
R.~Stevens, J.~Malone, S.~Williams, R.~Power, and A.~Third.
\newblock Automating generation of textual class definitions from {{OWL}} to {{English}}.
\newblock \emph{Journal of Biomedical Semantics}, 2\penalty0 (S2):\penalty0 S5, Dec. 2011.
\newblock ISSN 2041-1480.
\newblock \doi{10.1186/2041-1480-2-S2-S5}.

\bibitem[Tablan et~al.(2008)Tablan, Damljanovic, and Bontcheva]{kaufmannQuerixNaturalLanguage}
V.~Tablan, D.~Damljanovic, and K.~Bontcheva.
\newblock A natural language query interface to structured information.
\newblock In S.~Bechhofer, M.~Hauswirth, J.~Hoffmann, and M.~Koubarakis, editors, \emph{The Semantic Web: Research and Applications}, pages 361--375, Berlin, Heidelberg, 2008. Springer Berlin Heidelberg.
\newblock ISBN 978-3-540-68234-9.

\bibitem[Taffa and Usbeck(2023)]{TaffaU23}
T.~A. Taffa and R.~Usbeck.
\newblock Leveraging llms in scholarly knowledge graph question answering.
\newblock In D.~Banerjee, R.~Usbeck, N.~Mihindukulasooriya, G.~Singh, R.~Mutharaju, and P.~Kapanipathi, editors, \emph{Proc. of Scholarly {QALD} 2023 SemREC co-located with 22nd International Semantic Web Conference {ISWC}, Athens, Greece, November 6-10}, volume 3592 of \emph{{CEUR} Workshop Proc.} CEUR-WS.org, 2023.

\bibitem[Yang et~al.(2023)Yang, Teng, Dong, and Bo]{Yang_24}
S.~Yang, M.~Teng, X.~Dong, and F.~Bo.
\newblock {LLM}-{B}ased {SPARQL} generation with selected schema from large scale knowledge base.
\newblock In H.~Wang, X.~Han, M.~Liu, G.~Cheng, Y.~Liu, and N.~Zhang, editors, \emph{Knowledge Graph and Semantic Computing: Knowledge Graph Empowers Artificial General Intelligence}, pages 304--316, Singapore, 2023. Springer Nature Singapore.
\newblock ISBN 978-981-99-7224-1.

\bibitem[Zhou et~al.(2024)Zhou, Li, Liu, Xu, Liu, and Bao]{ZHOU2024102333}
B.~Zhou, X.~Li, T.~Liu, K.~Xu, W.~Liu, and J.~Bao.
\newblock Causal{KGPT}: {I}ndustrial structure causal knowledge-enhanced large language model for cause analysis of quality problems in aerospace product manufacturing.
\newblock \emph{Advanced Engineering Informatics}, 59:\penalty0 102333, 2024.
\newblock ISSN 1474-0346.

\bibitem[Zhu et~al.(2023)Zhu, Wang, Chen, Qiao, Ou, Yao, Deng, Chen, and Zhang]{Zhu-24}
Y.~Zhu, X.~Wang, J.~Chen, S.~Qiao, Y.~Ou, Y.~Yao, S.~Deng, H.~Chen, and N.~Zhang.
\newblock {LLM}s for {K}nowledge {G}raph construction and reasoning: Recent capabilities and future opportunities.
\newblock \emph{CoRR}, abs/2305.13168, 2023.
\newblock \doi{10.48550/ARXIV.2305.13168}.
\newblock URL \url{https://doi.org/10.48550/arXiv.2305.13168}.

\end{thebibliography}

\end{document}